\documentclass[11pt,a4paper]{article}
\usepackage[hyperref]{emnlp-ijcnlp-2019}
\usepackage{times}
\usepackage{latexsym}
\usepackage{amsmath}
\usepackage{mathrsfs}
\usepackage{tabularx}
\usepackage{graphicx}
\usepackage{amsthm}

\usepackage{url}

\usepackage[utf8]{inputenc}
\usepackage{booktabs}
\usepackage{tabularx}
\usepackage{bbm}
\usepackage{amsfonts}
\usepackage{multirow}
\usepackage{subfig}
\usepackage{natbib}
\usepackage{amsmath,bm}

\aclfinalcopy 

\title{Mitigating Annotation Artifacts in Natural Language Inference Datasets \\ to Improve Cross-dataset Generalization Ability}

\author{  
  Guanhua Zhang$^{1,2*}$, Bing Bai$^{1*}$, Junqi Zhang$^{1}$, Kun Bai$^1$,Conghui Zhu$^2$, Tiejun Zhao$^2$\\
  $^1$Cloud and Smart Industries Group, Tencent, China\\
  $^2$Harbin Institute of Technology, China\\
  \texttt{\{guanhzhang,icebai,benjqzhang,kunbai\}@tencent.com},\\
  \texttt{\{chzhu,tjzhao\}@hit-mtlab.net}
}
\date{}

\begin{document}
\maketitle
\begin{abstract}

Natural language inference~(NLI) aims at predicting the relationship between a given pair of premise and hypothesis. 
However, several works have found that there widely exists a bias pattern called annotation artifacts in NLI datasets, making it possible to identify the label only by looking at the hypothesis.
This irregularity makes the evaluation results over-estimated and affects models' generalization ability.
In this paper, we consider a more trust-worthy setting, \emph{i.e.}, cross-dataset evaluation.
We explore the impacts of annotation artifacts in cross-dataset testing.
Furthermore, we propose a training framework to mitigate the impacts of the bias pattern.
Experimental results demonstrate that our methods can alleviate the negative effect of the artifacts and improve the generalization ability of models.

\end{abstract}

\section{Introduction}
Natural language inference~(NLI) is a widely-studied problem in natural language processing. It aims at comparing a pair of sentences~(\emph{i.e.} a \emph{premise} and a \emph{hypothesis}), and inferring the relationship between them~(\emph{i.e.}, \texttt{entailment}, \texttt{neutral} and \texttt{contradiction}).
Large-scaled datasets like SNLI~\citep{bowman2015large} and MultiNLI~\citep{williams2017broad} have been created by crowd-sourcing and fertilized NLI research substantially.

However, several works~\citep{gururangan2018annotation, tsuchiya2018performance,wang2018if} have pointed out that crowd-sourcing workers have brought a bias pattern named \emph{annotation artifacts} in these NLI datasets.
Such artifacts in hypotheses can reveal the labels and make it possible to predict the labels solely by looking at the hypotheses.
For example, models trained on SNLI with only the hypotheses can achieve an accuracy of 67.0\%, despite the \emph{always predicting the majority-class} baseline is only 34.3\% ~\citep{gururangan2018annotation}. 

Classifiers trained on NLI datasets are supposed to make predictions by understanding the semantic relationships between given sentence pairs. 
However, it is shown that models are unintentionally utilizing the annotation artifacts~\citep{wang2018if,gururangan2018annotation}. If the evaluation is conducted under a similar distribution as the training data, \emph{e.g.}, with the given testing set, models will enjoy additional advantages, making the evaluation results over-estimated. On the other hand, if the bias pattern cannot be generalized to the real-world, it may introduce noise to models, thus hurting the generalization ability. 


In this paper, we use cross-dataset testing to better assess models' generalization ability.
We investigate the impacts of annotation artifacts in cross-dataset testing.
Furthermore, we propose an easy-adopting debiasing training framework, which doesn't require any additional data or annotations, and apply it to the high-performing Densely Interactive Inference Network~\citep[DIIN;][]{gong2017natural}.
Experiments show that our method can effectively mitigate the bias pattern and improve the cross-dataset generalization ability of models. 
To the best of our knowledge, our work is the first attempt to alleviate the annotation artifacts without any extra resources.

\section{Related Work}
\label{sec:related_work}
Frequently-used NLI datasets such as SNLI and MultiNLI are created by crowd-sourcing~\citep{bowman2015large, williams2017broad}, during which they present workers a premise and ask them to produce three hypotheses corresponding to labels.
As \citet{gururangan2018annotation} pointed out, workers may adopt some specific annotation strategies and heuristics when authoring hypotheses to save efforts, which produces certain patterns called annotation artifacts in the data.
Models' trained on such datasets are heavily affected by the bias pattern~\citep{gururangan2018annotation}.

\citet{wang2018if} further investigate models' robustness to the bias pattern using swapping operations. 
\citet{poliak2018hypothesis} demonstrate that the annotation artifacts widely exist among NLI datasets. 
They show that hypothesis-only-model, which refers to models trained and predict only with hypotheses, outperforms \emph{always predicting the majority-class} in six of ten NLI datasets.

The emergence of the pattern can be due to selection bias~\citep{rosenbaum1983central, zadrozny2004learning, d1998propensity} in the datasets preparing procedure.
Several works~\citep{levy2016annotating,rudinger2017social} investigate the bias problem in relation inference datasest.
\citet{zhang2019selection} investigate the selection bias embodied in the comparing relationships in six natural language sentence matching datasets and propose a debiasing training and evaluation framework.

\section{Making Artifacts Unpredictable}
\label{sec:our_method}
Essentially speaking, the problem of the bias pattern is that the artifacts in hypotheses are distributed differently among labels, so balancing them across labels may be a good solution to alleviate the impacts~\citep{gururangan2018annotation}. 

Based on the idea proposed by \citet{zhang2019selection}, we demonstrate that we can make artifacts in biased datasets balanced across different classes by assigning specific weights for every sample. 
We refer the distribution of the acquired weighted dataset as \emph{artifact-balanced distribution}.
We consider a supervised NLI task, which is to predict the relationship label $y$ given a sentence pair $x$, and we denote the hypothesis in $x$ as $h$. 
Without loss of generality, we assume that the prior probability of different labels is equal, and then we have the following theorem.

\newtheorem{theorem}{Theorem}
\begin{theorem}
\label{theorem:theorem}
For any classifier $f=f(x, h)$, and for any loss function $\Delta(f(x, h), y)$, if we use $w = \frac{1}{P(y|h)}$ as weight for every sample during training, it's equivalent to training with the artifact-balanced distribution.
\end{theorem}

Detailed assumptions and the proof of the theorem is presented in Appendix~\ref{sec:proof}.
With the theorem, we can simply use cross predictions to estimate $P(y|h)$ in origin datasets and use them as sample weights during training.
The step-by-step procedure for artifact-balanced learning is presented in Algorithm~1.

However, it is difficult to precisely estimate the probability $P(y|h)$.
A minor error might lead to a significant difference to the weight, especially when the probability is close to zero.
Thus, in practice, we use $w = \frac{1}{(1-\epsilon)P(y|h) + \epsilon}$ as sample weights during training in order to improve the robustness.
We can find that as $\epsilon$ increases, the weights tend to be uniform, indicating that the debiasing effect decreases as the smooth term grows.
Moreover, in order to keep the prior probability $P(Y)$ unchanged, we normalize the sum of weights of the three labels to the same.

\begin{table}[t!]
  \begin{center}
    \resizebox{0.485\textwidth}{!}{
      \begin{tabular}{p{0.15cm}<{\centering} p{8.55cm}}
        \hline
        \multicolumn{2}{l}{\textbf{Algorithm 1: Artifact-balanced Learning}}\\
        \hline
        \multicolumn{2}{p{9.2cm}}{\textbf{Input}: The dataset $\{x, h, y\}$ and the number of fold $K$ for cross prediction.} \\
        \multicolumn{2}{l}{\textbf{Procedure}:}\\
        01 & Estimate $P(y|h)$ for every sample by training classifiers and using $K$-fold cross-predicting strategy.\\
        02 & Obtain the weights $w=\frac{1}{(1-\epsilon)P(y|h)+\epsilon}$ for all samples and normalize the sum of the weights.\\
        03 & Train and validate models using $w$ as the sample weights.\\
        \hline
      \end{tabular}}
  \end{center}
\end{table}

\section{Experimental Results}
\label{sec:experimental_results}
\renewcommand\arraystretch{1.0}
\begin{table*}[t!]
\centering
\noindent\makebox[\textwidth]{
\resizebox{0.97\textwidth}{!}{
\begin{tabular}{c|c|c|c|c|c|c|c|c|c}
\hline
\multirow{3}*{\bf Trainset} & \multirow{3}*{\bf Model} & \multirow{3}*{\bf Smooth} & \multicolumn{5}{c|}{\bf Cross-dataset Testing} & \multicolumn{2}{c}{\bf Hard-Easy Testing} \\
\cline{4-10}
~ & ~ & ~ & \multicolumn{3}{c|}{\bf Human Elicited} & \multicolumn{2}{c|}{\bf Human Judged} & \multirow{2}*{\bf Hard} & \multirow{2}*{\bf Easy}  \\
\cline{4-8}
~ & ~ & ~ & {\bf SNLI} & {\bf MMatch} & {\bf MMismatch} & {\bf SICK} & {\bf JOCI} & ~ & ~ \\
\hline
\multirow{10}*{\bf SNLI} & \multirow{5}*{\bf Hyp} & 0.000 & 0.4776 & 0.4968 & 0.5013 & 0.4923 & 0.5016 & 0.5190 & 0.4587  \\
~ & ~ & 0.001 & 0.4779 & 0.4952 & 0.4924 & 0.4934 & 0.5044 & 0.5242 & 0.4568  \\
~ & ~ & \bf 0.010& \bf 0.5318 & \bf 0.5092 & \bf 0.5097 & \bf 0.5036 & \bf 0.4961 & \bf 0.5225 & \bf 0.5375 \\
~ & ~ & 0.100 & 0.7749 & 0.6009 & 0.6124 & 0.6060 & 0.5304& 0.5179 & 0.8811  \\
\cline{3-10}
~ & ~ & \bf  Baseline & \bf 0.8496 & \bf 0.6305 & \bf 0.6399 & \bf 0.6250 & \bf 0.5080 & \bf 0.4793 & \bf 0.9755  \\
\cline{2-10}
~ & \multirow{5}*{\bf Norm} & 0.000 & 76.61\% & 50.51\% & 51.50\% & 52.63\%$^*$ & 44.15\% & 74.36\%$^*$ & 77.72\% \\
~ & ~ & 0.001 & 72.75\% & 45.05\% & 46.24\% & 48.25\% & 39.68\% & 72.95\% & 72.65\%  \\
~ & ~ & \bf 0.010 & \bf 78.94\% & \bf 54.53\% & \bf 55.97\% & \bf 52.68\%$^*$ & \bf 46.19\% $^*$& \bf 75.38\%$^*$ & \bf 80.71\%  \\
~ & ~ & 0.100 & 83.57\% & 57.77\% & 60.37\% & 53.45\%$^*$ & 47.84\%$^*$ & 76.02\%$^*$ & 87.32\% \\
\cline{3-10}
~ & ~ & \bf Baseline & \bf 86.98\% & \bf 61.95\% & \bf 64.00\% & \bf 52.07\% & \bf 45.63\% & \bf 73.81\% & \bf 93.52\%  \\
\hline
\multirow{12}*{\bf MultiNLI} & \multirow{6}*{\bf Hyp} & 0.000 & 0.4647 & 0.4427 & 0.4429 & 0.4685 & 0.4874 & 0.4957 & 0.3998 \\
~ & ~ & 0.001 & 0.4433 & 0.4174 & 0.4152 & 0.4583 & 0.4933 & 0.4969 & 0.3534  \\
~ & ~ & 0.010 & 0.4560 & 0.4562 & 0.4590 & 0.4723 & 0.4970 & 0.4992 & 0.4201  \\
~ & ~ & \textbf{0.020} & \textbf{0.4741} & \textbf{0.4850} & \textbf{0.4957} & \textbf{0.5003} & \textbf{0.4969} & \textbf{0.5006} & \textbf{0.4703}  \\
~ & ~ & 0.100 & 0.5711 & 0.6482 & 0.6596 & 0.5944 & 0.5208 & 0.5023 & 0.7619  \\
\cline{3-10}
~ & ~ & \bf Baseline & \bf 0.6483 & \bf 0.7252 & \bf 0.7253 & \bf 0.6079 & \bf 0.4587 & \bf 0.4998 & \bf 0.8915  \\
\cline{2-10}
~ & \multirow{6}*{\bf Norm} & 0.000 & 52.06\% & 58.92\% & 60.63\% & 52.99\% & 48.27\%$^*$ & 56.80\% & 60.78\%  \\
~ & ~ & 0.001 & 53.90\% & 59.48\% & 60.50\% & 52.70\% & 45.67\%$^*$ & 58.19\% & 60.61\%  \\
~ & ~ & 0.010 & 58.13\% & 62.82\% & 64.35\% & 54.17\% & 45.78\%$^*$ & 61.27\% & 64.18\%  \\
~ & ~ & \bf 0.020 & \bf 61.37\% & \bf 66.68\% & \bf 68.18\% & \bf 57.20\%$^*$  & \bf 48.59\%$^*$ & \bf 62.21\% & \bf 70.60\% \\
~ & ~ & 0.100 & 64.16\% & 71.54\% & 72.77\% & 58.35\%$^*$ & 48.81\%$^*$ & 66.14\% & 76.28\%  \\
\cline{3-10}
~ & ~ & \bf Baseline & \bf 68.49\% & \bf 76.20\% & \bf 76.38\% & \bf 56.74\% & \bf 41.18\% & \bf 66.24\% & \bf 84.92\%  \\
\hline
\end{tabular}}}
\caption{\label{table:result} Evaluation Results of \emph{Hyp} and \emph{Norm}. 
\emph{Baseline} refers to the model trained and validated without using weights. 
\emph{Hard}, \emph{Easy} refers to the \emph{Hard-Easy Testing} generated from the testing set corresponding to the \emph{Trainset} column.
Results of \emph{Hyp} are the average numbers of five runs with different random initialization. 
{\bf We report AUC for \emph{Hyp} and ACC for \emph{Norm}.}
``*'' indicates where normal-model are better than the baseline.}
\end{table*}
In this section, we present the experimental results for cross-dataset testing of artifacts and artifact-balanced learning.
We show that cross-dataset testing is less affected by annotation artifacts, while there are still some influences more or less in different datasets. 
We also demonstrate that our proposed framework can mitigate the bias and improve the generalization ability of models.

\subsection{Evaluation Scheme}
\paragraph{Cross-dataset Testing} We utilize SNLI \citep{bowman2015large}, MultiNLI \citep{williams2017broad}, JOCI \citep{zhang2017ordinal} and SICK \citep{marelli2014semeval} for cross-dataset testing.

SNLI and MultiNLI are prepared by \emph{Human Elicited}, in which workers are given a context and asked to produce hypotheses corresponding to labels.
SICK and JOCI are created by \emph{Human Judged}, referring that hypotheses and premises are automatically paired while labels are generated by humans~\citep{poliak2018hypothesis}. 
In order to maximumly mitigate the impacts of annotation artifacts during evaluations, we train and validate models respectively on SNLI and MultiNLI and test on both SICK and JOCI.
We also report models' performances on SNLI and MultiNLI.

As to SNLI, we use the same partition as \citet{bowman2015large}. 
For MultiNLI, we separately use two origin validation sets~(\emph{Matched} and \emph{Mismatched}) as the testing sets for convenience, and refer them as \emph{MMatch} and \emph{MMismatch}. 
We randomly select 10000 samples out of the origin training set for validation and use the rest for training.
As to JOCI, we use the whole ``B'' subsets for testing, whose premises are from SNLI-train while hypotheses are generated based on world knowledge~\citep{zhang2017ordinal}, and convert the score to NLI labels following \citet{poliak2018hypothesis}.
As to SICK, we use the whole dataset for testing.

\paragraph{Hard-Easy Testing}  To determine how biased the models are, we partition the testing set of SNLI and \emph{MMatch} into two subsets: examples that the hypothesis-only model can be correctly classified as \emph{Easy} and the rest as \emph{Hard} as seen in \citet{gururangan2018annotation}. 
More detailed information is presented in Appendix~\ref{app:hard_easy}.

\subsection{Experiment Setup}
We refer models trained only with hypotheses as \emph{hypothesis-only-model} (\emph{Hyp}), and models that utilize both premises and hypotheses as \emph{normal-model} (\emph{Norm}). 
We implement a simple LSTM model for \emph{Hyp} and use DIIN~\cite{gong2017natural}
\footnote{https://github.com/YichenGong/Densely-Interactive-Inference-Network}
as \emph{Norm}.
We report AUC\footnote{We calculate metrics for each label and report their mean.} for \emph{Hyp} and ACC for \emph{Norm}.
More details can be seen in Appendix~\ref{sec:setting}

We estimate $P(y|h)$ for SNLI and MultiNLI respectively using BERT~\citep{devlin2018bert} with 10-fold predictions. 
To investigate the impacts of smooth terms, we choose a series of smooth values and present the results.
Considering models may jiggle during the training phase due to the varied scale of weights, we sample examples with probabilities proportional to the weights for every mini-batch instead of adding weights to the loss directly.

The evaluation results are reported in Table~\ref{table:result}.

\subsection {Can Artifacts Generalize Across Datasets?}

\textbf{Anotation Artifacts can be generalized across Human Elicited datasets.} From the AUC of \emph{Hyp} baseline trained with SNLI, we can see that the bias pattern of SNLI has a strong predictive ability in itself and the other two testing sets of \emph{Human Elicited}. The behavior of those trained with MultiNLI is similar.

\textbf{Anotation Artifacts of SNLI and MultiNLI can be generalized to SICK.} Unexpectedly, it is shown that \emph{Hyp} baseline can get $0.6250$~(AUC) trained with SNLI and $0.6079$~(AUC) with MultiNLI when tested on SICK, indicating that the bias pattern of SNLI and MultiNLI are predictive on SICK.
The results imply that the bias pattern can even be generalized across datasets prepared by different methods.

\textbf{Annotation Artifacts of SNLI are nearly neutral in JOCI, while MultiNLI is misleading.}
We find that AUC of \emph{Hyp} baseline trained with SNLI is very close to $0.5$ on JOCI, indicating that JOCI is nearly neutral to artifacts in SNLI.
However, when it comes to training with MultiNLI, the AUC of \emph{Hyp} baseline is lower than $0.5$, indicating that the artifacts are misleading in JOCI.

\subsection {Debiasing Results}
\paragraph{{Effectiveness of Debiasing}}
Focusing on the results when smooth equals $0.01$ for SNLI and smooth equals $0.02$ for MultiNLI, we observe that the AUC of \emph{Hyp} for all testing sets are approximately $0.5$, indicating \emph{Hyp}'s predictions are approximately equivalent to randomly guessing. 
Also, the gap between \emph{Hard} and \emph{Easy} for \emph{Norm} significantly decreases comparing with the baseline. 
With the smooth, we can conclude that our method effectively alleviates the bias pattern.

With other smooth terms, our method still has more or less debiasing abilities.
In those testing sets which are not neutral to the bias pattern, the AUC of \emph{Hyp} always come closer to $0.5$ comparing with the baseline with whatever smooth values.
Performances of \emph{Norm} on \emph{Hard} and \emph{Easy} also come closer comparing with the baseline.
\emph{Norm} trained with SNLI even exceed baseline in \emph{Hard} with most smooth terms.

From the results of \emph{Hyp}, we can find a trend that the larger the smooth value is, the lower the level of debiasing is, while with a very small or even no smooth value, the AUC may be lower than $0.5$. 
As mentioned before, we owe this to the imperfect estimation of $P(y|h)$, and we can conclude that a proper smooth value is a prerequisite for the best debiasing effect.

\paragraph{Benefits of Debiasing}
Debiasing may improve models' generalization ability from two aspects: (1)~Mitigate the misleading effect of annotation artifacts. (2)~Improve models' semantic learning ability.

When the annotation artifacts of the training set cannot be generalized to the testing set, which should be more common in the real-world, predicting by artifacts may hurt models' performance.
Centering on the results of JOCI, in which the bias pattern of MultiNLI is misleading, we find that \emph{Norm} trained with MultiNLI outperforms baseline after debiasing with all smooth values tested. 

Furthermore, debiasing can reduce models' dependence on the bias pattern during training, thus force models to better learn semantic information to make predictions.
\emph{Norm} trained with SNLI exceed baseline in JOCI with smooth terms $0.01$ and $0.1$. 
With larger smooth terms, \emph{Norm} trained with both SNLI and MultiNLI exceeds baseline in SICK.
Given the fact that JOCI is almost neutral to artifacts in SNLI, and the bias pattern of both SNLI and MultiNLI are even predictive in SICK, we owe these promotions to that our method improves models' semantic learning ability.

As to other testing sets like SNLI, \emph{MMatch} and \emph{MMismatch}, we notice that the performance of \emph{Norm} always decreases compared with the baseline. 
As mentioned before, both SNLI and MultiNLI are prepared by \emph{Huamn Elicited}, and their artifacts can be generalized across each other.
We owe the drop to that the detrimental effect of mitigating the predictable bias pattern exceeds the beneficial effect of the improvement of semantic learning ability.

\section{Conclusion}
\label{sec:conclusion}
In this paper, we take a close look at the annotation artifacts in NLI datasets. 
We find that the bias pattern could be predictive or misleading in cross-dataset testing. 
Furthermore, we propose a debiasing framework and experiments demonstrate that it can effectively mitigate the impacts of the bias pattern and improve the cross-dataset generalization ability of models.
However, \textbf{it remains an open problem that how we should treat the annotation artifacts}. We cannot assert whether the bias pattern should not exist at all or it is actually some kind of nature. We hope that our findings will encourage more explorations on reliable evaluation protocols for NLI models.

\clearpage
\bibliography{emnlp-ijcnlp-2019}
\bibliographystyle{acl_natbib}
\clearpage
\appendix
\section{Detailed Assumptions and Proof of Theorem~\ref{theorem:theorem}}
\label{sec:proof}
We make a few assumptions about an artifact-balanced distribution and how the biased datasets are generated from it, and demonstrate that we can train models fitting the artifact-balanced distribution using only the biased datasets.

We consider the domain of the artifact-balanced distribution $\mathscr{D}$ as $\mathcal{X} \times \mathcal{A} \times \mathcal{Y} \times \mathcal{S}$, in which $\mathcal{X}$ is the input variable space, $\mathcal{Y}$ is the label space, $\mathcal{A}$ is the feature space of annotation artifacts in hypotheses, $\mathcal{S}$ is the selection intention space.
We assume that the biased distribution $\widehat{\mathscr{D}}$ of origin datasets can be generated from the artifact-balanced distribution by selecting samples with $S = Y$, i.e., the selection intention matches with the label.
We use $P(\cdot)$ to represent the probability on $\widehat{\mathscr{D}}$ and use $Q(\cdot)$ for $\mathscr{D}$.

We also make some assumptions about the artifact-balanced distribution.
The first one is that the label is independent with the artifact in the hypothesis, defined as follows,
\begin{displaymath}
  Q(Y|A) = Q(Y) \text{.}
\end{displaymath}
The second one is that the selection intention is independent with $X$ and $Y$ when the annotation artifact is given,
\begin{displaymath}
  Q(S|X, A, Y) = Q(S|A) \text{.}
\end{displaymath}

And we can prove the equivalence of training with weight $\frac{1}{P(Y|A)}$ and fitting the artifact-balanced distribution.
We first present an equation as follows,
\begin{equation*}  
\begin{split}
  &P(Y=0|A) \\
 =&  Q(Y=0|A,S=Y)\\
 =&  \frac{Q(S=Y=0|A)}{Q(S=Y|A)} \\
 =&  \frac{Q(S=0|A)Q(Y=0|A)}{\sum_{i=0}^{2} Q(S=i|A)Q(Y=i|A)} \\
 =&  \frac{Q(S=0|A)Q(Y=0)}{\sum_{i=0}^{2} Q(S=i|A)Q(Y=i)} \\
\end{split}
\end{equation*}
Without loss of generality, we can assume $Q(Y=i)=\frac{1}{3}~(i=0,1,2)$ and get that,
\begin{displaymath}
  P(Y=i|A) = Q(S=i|A)~~~~(i=0,1,2)\text{.}
\end{displaymath}
With the above derivation, we can prove the equivalence like following,
\begin{proof}
\begin{small}
\begin{displaymath}
\begin{split}
& E_{x,a,y \sim \widehat{\mathscr{D}}} \Big[\frac{1}{P(Y=y|a)}\Delta\big(f(x,a), y\big) \Big] \\
= & \int \Delta(f(x,a), y) \frac{1}{P(Y=y|a)} dP(x,a,y) \\
= & \int \Delta(f(x,a), y) \frac{1}{Q(S=y|a)} dQ(x,a,y|S=Y) \\
= & \int \Delta(f(x,a), y) \frac{1}{Q(S=y|a)} \frac{Q(S=y|x,a,y)dQ(x,a,y)}{Q(S=Y)} \\
= & \int \Delta(f(x,a), y) \frac{1}{Q(S=y|a)} \frac{Q(S=y|a)}{Q(S=Y)} dQ(x,a,y)\\
= & \frac{1}{Q(S=Y)} E_{x,a,y \sim \mathscr{D}} \Big[\Delta\big(f(x,a), y\big) \Big]
\end{split}
\end{displaymath}
\end{small}
\end{proof}
As $Q(S=Y)$ is just a constant, training with the loss is equivalent to fitting the artifact-balanced distribution. 
Given hypotheses variable \emph{H}, the probability $P(Y|A)$ can be replaced by $P(Y|H)$ since the predictive ability of hypotheses totally comes from the annotation artifacts, and we can have $w=\frac{1}{P(Y|H)}$ as weights during training.

\section{Experiment Setting}
\subsection{Hard-Easy Datasets Setting}
\label{app:hard_easy}
For SNLI, we use \emph{Hard} released by \citet{gururangan2018annotation}.
For \emph{MMatch}, we manually partition the set using fastText~\citep{joulin2016bag}.
And we summarize the size of the datasets used in \emph{Hard-Easy Testing} below.

\begin{center}
\begin{tabular}{c|c|c}
\hline
\bf Trainset & \bf Hard & \bf Easy  \\ 
\hline
SNLI        & 3261 & 6563 \\
\hline
MultiNLI    & 4583 & 5232 \\
\hline
\end{tabular}
\end{center}

\subsection{Experiment Setup}
\label{sec:setting}
For DIIN, we use settings same as \citet{gong2017natural} but do not use syntactical features. Priors of labels are normalized to be the same.
For hypothesis-only-model, we implement a na\"ive model with one LSTM layer and a three-layer MLP behind, implemented with Keras and Tensorflow backend~\citep{abadi2016tensorflow}.
We use the 300-dimensional GloVe embeddings trained on the Common Crawl 840B tokens dataset~\citep{pennington2014glove} and keep them fixed during training. 
Batch Normalization~\citep{ioffe2015batch} are applied after every hidden layer in MLP and we use Dropout~\citep{srivastava2014dropout} with rate 0.5 after the last hidden layer.
We use RMSProp\citep{tieleman2012lecture} as optimizer and set the learning rate as 1e-3.
We set the gradient clipping to 1.0 and the batch size to 256. 

\end{document}